# On Some Integrated Approaches to Inference

Mark A. Kon[1], Boston University

Leszek Plaskota[2], University of Warsaw

**Abstract:** We present arguments for the formulation of unified approach to different standard continuous inference methods from partial information. It is claimed that an explicit partition of information into *a priori* (prior knowledge) and *a posteriori* information (data) is an important way of standardizing inference approaches so that they can be compared on a normative scale, and so that notions of optimal algorithms become farther-reaching. The inference methods considered include neural network approaches, information-based complexity, and Monte Carlo, spline, and regularization methods. The model is an extension of currently used continuous complexity models, with a class of algorithms in the form of optimization methods, in which an optimization functional (involving the data) is minimized. This extends the family of current approaches in continuous complexity theory, which include the use of interpolatory algorithms in worst and average case settings.

## 1. Introduction

In this paper we extend and extrapolate some elements of the theory optimal algorithms ([TW]) and continuous complexity theory [TWW], so as to include and compare larger classes of continuous algorithms. The theory of function approximation has over a number of years led to many new and integrated approaches in statistics, statistical learning theory, neural network theory, and related fields. Indeed, there are a large number of areas of mathematics, statistics, and computer science which deal with extrapolation of functions from partial information or examples.

The problems in these areas can be summarized in the function approximation problem: How can we best estimate the function $f$ from partial information (examples) $y = Nf = (f(x_1) + \epsilon_1, \ldots, f(x_k) + \epsilon_k)$ consisting of the values of $f$ at a finite number of points, with possible error $\epsilon_k$? Put somewhat more broadly, given a normed linear space $F$ and an unknown $f \in F$, how can we best estimate $f$ (in the norm of $F$) if we have information $Nf = (L_1 f, \ldots, L_k f)$, where $L_i$ are (linear or nonlinear) functionals (we henceforth implicitly assume possible presence of error terms $\epsilon_k$ in the components

---

[1]Research partially supported by the National Science Foundation and the U.S. Fulbright Commission
[2]Research partially supported by the Polish-American Fulbright Foundation and the National Committee for Scientific Research of Poland

of $N$). In [TW1] and [TWW] (see also [TW2]), a theory of information and algorithmic complexity has been developed in the study of the function approximation problem and its generalizations.

For an input-output (i-o) function $f(x)$, the input $x$ effectively codes a problem we wish to solve (e.g., a visual field light intensity function), and the output $f(x)$ codes a solution (e.g., identification of object pictured). The function $f$ is generally not known explicitly, and only partial information (in the form of examples) $y = Nf = (f(x_1), f(x_2), \ldots, f(x_n))$ is given about $f$. The goal, as mentioned above, is to identify $f$ from this partial information.

Work dealing with the function approximation problem is closely related to statistical learning theory [Va]. Other theories and approaches have included learning in neural network theory [RHW], computational learning theory [KV], regularization theory [Ti, TA, PG1,2], regression theory in statistics, the maximum entropy method [Ja1,2], the theory of V-C dimension and approximation [Va, PG2], and approximation theory [MR]. Some of these theories are much more inclusive than others; we wish to develop a theory of optimal algorithms in the broader context. We also wish more precisely to define optimality within the classes of approaches, and to be able to identify optimal algorithms within them. From this we can form a normative index of such methods according to their optimality properties.

In this paper we attempt to integrate a number of approaches to the function approximation problem, in order to expand the basis for comparison of methods and algorithms. A number of currently used methods seem outside of the domain of the information-based continuous complexity model, but we will show that a specific class of algorithms in this model includes these existing approaches. This will hopefully move us closer to formulation of a more inclusive theory of continuous optimal algorithms, one into which most current approaches for function extrapolation would fit.

In particular we argue for a unified approach to the problem of function approximation, which has been studied through large numbers of different approaches, through an explicit separation of information into *a priori* and *a posteriori* information. The first, *a priori information,* consists of our prior information regarding the i-o function $f$ we seek to approximate, before data are gathered. *A posteriori* information consists of data $Nf$. We believe almost all approaches to prediction and classification can be formulated in a uniform setting classifying methods of combining a priori and a posteriori information, as is done in information-based complexity.

In most continuous complexity approaches, *a priori* information usually indicates that $f$ belongs to a balanced convex set of functions $F_1$ (the *a priori* set) in a normed linear space $F$. The *a posteriori* information $y = Nf$ restricts the class of potential functions $f$ to be in $N^{-1}y$. Optimal algorithmic solutions to the problem of estimating $f$ in the norm $\|\cdot\|_F$ consist of finding the center (or approximating it for almost optimal solutions) of the set $F_1 \cap N^{-1}y$ through some algorithm $\phi$, and $\phi(Nf)$ is the approximation to $f$ obtained through this procedure. We have an *a priori* class $F_1$ and an *a posteriori* class $N^{-1}y$, and we seek the "true" $f$ in their intersection. We will call this the *interpolatory approach.*

In maximum likelihood methods of Bayesian statistics, the approach is generally an optimization approach, dealing with *a priori* optimization functionals (i.e., the *a priori* probability distributions), which are to be optimized to be consistent with *a posteriori* data $Nf$. On the other hand, regression approaches in statistics have a more implicit partition of a priori and a posteriori information. Such a division can be made explicit, and these approaches also can be classified in the optimization part of our model.

Our approach extends the above (involving algorithms $\phi$ which interpolate in $F_1 \cap N^{-1}y$) to include optimization algorithms such as those in maximum likelihood mentioned above. In general, an optimization algorithm $\phi$ can have one of two forms. First, it can approximate a desired $f^* \in N^{-1}y \cap F_1$ which minimizes a functional $\Lambda(f)$ subject to the constraint $f \in N^{-1}y \cap F_1$. Here the functional $\Lambda(\lambda)$ incorporates *a priori* preferences for functions with lower values of $\lambda$. Second, if error terms $\epsilon_i$ affect the measured values $f(x_i)$ in $Nf$, such an algorithm can compromise between the a posteriori requirement $Nf = y$ (with $y$ the data) and the a priori one that $\Lambda(f)$ be small. This can be done via minimization of a weighted combination $H_\lambda(f) = \|Nf - y\|^2 + \lambda\Lambda(f)$. This approach will be called the *optimization approach.*

In fact, we feel that the union of the interpolatory and optimization approaches constitutes a very comprehensive set of algorithms $\phi(Nf)$ which explicitly separate the two types of information.

The approaches we discuss here should be compared with the so-called V-C approaches of Vapnik and Chervonenkis [Va, VC]), in which a standardization is invoked through the construction of indexed familles $\{V_\lambda\}$ of candidate *a priori* spaces increasing in complexity with $\lambda$. There the approach is to find a $\lambda$ sufficiently small that the candidate set $N^{-1}y \cap V_\lambda$ is small, so there is a reasonable process of selecting the approximation $f^*$ of $f$ from $N^{-1}y \cap V_\lambda$. If this process is taken to the point where we find the *smallest* $\lambda$ so that $N^{-1}y \cap V_\lambda$ is still non-empty (assumed for the moment to exist), then this set will consist of those points $f^* \in N^{-1}y$ (i.e., consistent with the data) which minimize the functional $H(f) = \inf\{\lambda: f \in V_\lambda\}$. Therefore, the algorithms we present which consist of optimization approaches are consistent with the V-C approach.

Neural network algorithms such as backpropagation and the Boltzmann machine use examples $Nf$ of an unknown i-o function $f$. Then they effectively apply an algorithm $\phi$ to $Nf$, computing from $f$ an approximation $f^* = \phi(Nf)$ chosen from a given parameterized class $P$ of network-computable functions. Let us consider the set $P$ of such functions (i-o functions for a network of fixed size, parameterized by its weights $w_i$) from which the "closest match" is sought. First, if $P$ is too large compared to the cardinality of the information $Nf$, (i.e., there are too many neurons in the network compared to examples $f(x_i)$), then there are too many different close matches, and the problem becomes ill-posed. If the class $P$ is not too large, then the choice of approximation $f^* \in P$ reflects the a priori assumption that the function $f$ can be closely approximated by an element of $P$. Such an *a priori* class $P$, essentially, consists of a sufficiently diverse class of smooth functions. Philosophically, this class is not much different from, say, a class $F$ consisting of a ball in a Sobolev space, i.e. an $F$ optimizing for smoothness. Indeed, when $P$ is specialized to be a class of radial basis functions, we know that optimal approximations from $P$ quite explicitly minimize for Sobolev norm. There are in fact many variations on this theme of optimization of i-o functions with implicit a priori assumptions essentially consisting of smoothness in the interpolation literature, formulated in different ways.

The adaptive resonance theory (ART) algorithm [Ca,CG] in its simplest (winner take all) form consists of a dynamic allocation procedure in which network weights are determined in such a way that for various (appropriately sized) regions $R_i$ in the input space $R$, different neurons $y_i$ in the second (recognition) layer will respond. Once the programming of such a network is complete, the network is similar (again, in its simplest from) to a feedforward RBF network, in which there is a second competitive processing stage where the hidden neuron with highest activation suppresses all other neurons. Effectively, this ART network computes the function

$$f^*(x) = \arg\sup(G_1(x), \ldots, G_n(x)), \tag{1}$$

where the right hand side represents the choice of the function $G_i(x)$ with the maximum value given the input $x$. Here $G_i(x)$ are the activation functions of the neurons in the second layer of the ART network. The choice of $G_i(x)$ might be coded as the choice of neuron in the second layer, $x_i$ if $G_i$ is a radial function centered at $x_i$.

We remark that the above i-o functions $f^*(x)$ are a new class from the standpoint of RBF networks, since they can be highly discontinuous functions, say of a type needed in vision algorithms. In the context an ART network built to approximate a single i-o function $f$, say the characteristic function of a single category which we wish the network to identify, the division into a priori and a posteriori information is again clear. *A posteriori* information is the data vector $x$, while *a priori* information consists of the fact that the i-o function will be approximable in the form (1), in this case with only two choices $G_1(x)$ (which represents membership in the category) and $G_2(x)$, which represents non-membership). This is equivalent to the assumption that the i-o function will be in the class of threshold functions of a difference $G_1(x) - G_2(x)$. This is in a sense an a priori assumption regarding smoothness of the separating classes $R_i$ above, placing the set of potential choices $R_i$ into an *a priori* set of partitions defined as above by the class of activation functions $G_i(x)$. The algorithm $\phi$ used on the information $Nf$ is a complex iterative one with the goal of finding the best approximation to $f(x)$ in the parametric family of functions $f^*$ above.

Finally, we note that viewpoints on learning from partial information have very close parallels to data compression theory. In data compression one transforms data into minimal form, and then uses a procedure (the decompression algorithm) on the minimal information in the compressed data to reproduce the original. Compression approaches effectively search for minimal information ways of coding data, and thus implicitly address the question of extrapolation of full data sets from this such minimal information.

This can be easily seen in the wavelet reconstruction algorithms for images developed by Mallat [Ma]. In this case the minimal information kept about an image to be compressed consists of the zeros of its wavelet transform. The decompression process takes these minimal (*a posteriori*) data together with some *a priori* information. The latter consists of the fact that the function to be recovered is the class of continuous wavelet transform. This exemplifies how algorithms which use minimal amounts of *a posteriori* information must rely on large amounts of *a priori* information regarding the object to be reconstructed. The iterated projection algorithm used by Mallat [Ma] is a good example of how unexpectedly effective algorithms interpolating a priori and a posteriori information can be constructed.

The Mallat algorithm is a good illustration of the principle that inference and compression are closely related. If we wish to be able to reconstruct from minimal information an element $f$ known to be in a set $F$, we might minimize information about $f$ by compressing it to $Nf$, where $N$ is a linear or nonlinear operator. The decompression from $y = Nf$ to $f$ is a regularization procedure. It is based on the fact that we know *a priori* that $f \in F$, and that the intersection $N^{-1}y \cap F$ is sufficiently well-defined (of small enough deterministic or average-case radius). Sufficiently well-defined means that its center or some point in it is a good approximation to $f$ itself. This of course contrasts with the set $N^{-1}y$, which is generally large, especially in the case of high compression ratios. This approach can be formulated in the language of optimization as well (see below), as interpolation approaches such as the above in fact can be incorporated into the class of optimization approaches.

## 2. Problems and solution strategies.

In the background of our general approach lies the assumption that the problem we want to solve can be described in terms of a mapping

$$S : F \to G.$$

Here $F$ is the set of all problem instances, and $G$ is the set of all possible solutions. For $f \in F$, the solution is given as $g = S(f)$. We do not assume anything special about the sets $F$ and $G$ right now. They can be discrete or have continuous character. For instance, if we want to know if a patient is ill at a particular moment, then $F = P \times T$ where $f$ is in the set of all potential patients and $T$ is a time interval, $G = \{yes, no\}$, and $S(p, t) = yes$ if and only if $p$ is ill at time $t$. This is a *decision problem*. In another example, suppose that one wants to compute the integral $\int_0^1 f(x)dx$. Then $F$ is a set of possible integrands $f : [0, 1] \to \mathbb{R}$, and $G = \mathbb{R}$. This is a *continuous problem.*

The problem is solved by means of an *algorithm*. The algorithm usually does not use $f$ itself as data, but only some *information about $f$*. For instance, a judgment if a patient is ill is made based on a finite number of parameters, such as body temperature, age, symptoms, etc. The reason for that is that it is impossible to know everything about the patient, but also that some characteristics can be simply neglected, as they would not change the answer. Similarly, if the integrand $f$ is a complicated function, one usually uses a quadrature $\sum_{j=1}^{n} a_j f(x_j)$ to approximate the integral. This quadrature obviously does not use $f$ itself, but only its values at a finite number of points.

Information about $f$ will be formally described as $y = N(f)$ where $N$ is an *information operator*,

$$N : F \to Y,$$

or more generally, $y = N(f, \mathfrak{n})$ where

$$N : F \times \mathfrak{N} \to Y.$$

The first case represents *deterministic* situation where the value $y$ of information depends on $f$ only. The second case is *non deterministic*, i.e., $y$ is also a function of a side parameter $\mathfrak{n} \in \mathfrak{N}$, which is usually *random*. This randomness may have different sources. In many cases, it is a result of *noise* in observations or computations, but also may represent randomness of the information itself. For instance, if $f$ is a function then we may have

$$y = N(f) = \big(f(x_1), f(x_2), \ldots, f(x_n)\big) \tag{1}$$

where $x_i$'s are some *sample points* from the domain of $f$ (exact information), or

$$y = N(f, \mathfrak{n}) = \big(f(x_1) + \mathfrak{n}_1, \ldots, f(x_n) + \mathfrak{n}_n\big) \tag{2}$$

where $\omega = (\omega_1, \omega_2, \ldots, \omega_n)$ represents noise in sampling (noisy information), or

$$y = N(f, \mathfrak{n}) = \Big(f\big(x_1(\mathfrak{n}_1)\big), f\big(x_2(\mathfrak{n}_2)\big), \ldots, f\big(x_n(\mathfrak{n}_n)\big)\Big) \tag{3}$$

where the points $x_i$'s are selected randomly (random or Monte Carlo information). (Obviously, we may also have Monte Carlo information with noise.)

Similarly, the algorithm using information $y \in Y$ is formally a mapping

$$A : Y \to G,$$

or more generally,

$$A : Y \times \mathfrak{A} \to G$$

with a set $\mathfrak{A}$ of random parameters. Thus, for $f \in F$, the algorithm $A$ produces the value $A(y, \mathfrak{a})$, where $y$ is information about $f$. Composing the information $N$ and the algorithm $A$ we obtain that the approximation $a$ to $g = S(f)$ is given as $a = A\big(N(f, \mathfrak{n}), \mathfrak{a}\big)$. In the pure deterministic case we just have $a = A(N(f))$.

**Remark.** We distinguished the deterministic and non deterministic cases for reader's convenience only. Deterministic information and algorithm are obtained by putting $\mathfrak{N}$ and $\mathfrak{A}$ as singletons.

A *strategy* is a single information and algorithm using it, $U = (A, N)$, or a collection $U = \{U_i\}_{i \in I}$, $U_i = (A_i, N_i)$, where $I$ is an index set. The purpose of introducing strategies is to group procedures that use the same idea for constructing approximate solutions.

**Example 1** *(Numerical integration)* The problem is to evaluate

$$S(f) = \int_0^1 f(x)\,dx$$

for a continuous function $f$. That is, $f \in F = C([0, 1])$ and $G = \mathbb{R}$. The information is assumed to be of the form (1) with arbitrary sample points $x_i$. Possible approximations are provided, e.g., by *trapezoidal rule*

$$U_n^T(f) = \frac{1}{2n}\left(f(0) + f(1) + 2\sum_{i=1}^{n-1} f\left(\frac{i}{n}\right)\right),$$

or by *Simpson quadratures*

$$U_n^S(f) = \frac{1}{6n}\left(f(0) + f(1) + 4\sum_{i=1}^{n} f\left(\frac{i - 1/2}{n}\right) + 2\sum_{i=1}^{n-1} f\left(\frac{i}{n}\right)\right).$$

We can think of strategies as single approximations $U_n^T$ or $U_n^S$, but it is more natural to consider only two different strategies, $U^T = \{U_n^T\}_{n \geq 1}$ and $U^S = \{U_n^S\}_{n \geq 1}$, since for all $n$ the approximations are in both cases constructed based on the same rules - one takes the integral of piecewise linear or piecewise quadratic function interpolating $f$. There are obviously many other strategies for approximate computation of integrals including, e.g., *Gauss-Legendre* quadratures. In the case of *Monte Carlo* quadratures (which are applied to multi-dimentional rather than one-dimensional integration), we use random information (3) with the random parameter $\mathfrak{n}$ uniformly distributed on $[0, 1]^n$, and

$$U_n^{MC}(f, \mathfrak{n}) = \frac{1}{n}\sum_{i=1}^{n} f\big(x_i(\mathfrak{n})\big).$$

This is another kind of strategy, $U^{MC} = \{U_n^{MC}\}_{n \geq 1}$.

**Example 2** *(Data fitting)* Suppose now we want to recover a surface from noisy data about its values at some points. That is, $F = G$ is a set of possible surfaces which are represented as functions $f: D \to \mathbb{R}$ with $D \subset \mathbb{R}^2$. The information $y$ is given by (2). We

write for convenience

$$y = N(f) + \mathfrak{n},$$

where $N(f) = \big(f(x_1), \ldots, f(x_n)\big)$ and $\mathfrak{n} \in \mathbb{R}^n$ is the *noise*. A standard strategy used in Bayesian statistics is to approximate the surface $f$ by a surface $g_y$ which minimizes the *penalty functional*

$$||g||^2 \;+\; \frac{\lambda}{n} \cdot \sum_{i=1}^{n} \big(y_i - g(x_i)\big)^2.$$

Here $||\cdot||$ is a norm on a subspace $F_0 \subset F$, the minimization is over $F_0$, and $\lambda$ is a suitably chosen positive parameter.

**Example 3** *(Assigning probabilities)* In this problem, we want to assign unknown probabilities $p_i$ to $m$ different events, based on information about expectations of some random variables. Formally, we are given a discrete set $\Omega = \{\omega_1, \omega_2, \ldots, \omega_m\}$, and $F = G$ are all possible probability distributions $f = \{f_1, f_2, \ldots, f_m\}$ on $\Omega$. We obviously assume $p_i \geq 0$ and $\sum_{i=1}^{m} p_i = 1$. The problem is to recover a distribution $f \in F$ from information $y = (y_1, \ldots, y_n)$, where

$$y = Mf$$

and $M$ is an $n \times m$ matrix, $n < m$. Note that, given $y$, we know for sure that $f$ is in the set $F(y) = F \cap M^{-1}(y)$. What element of $F(y)$ should we choose as an approximate solution? There are many possible strategies. For instance, we can take as $U(f)$ the *(Chebyshev) center* of $F(y)$ with respect to, say, Euclidean norm in $\mathbb{R}^m$, or an element with minimal uniform norm. Physicists, however, prefer to choose the distribution $f \in F(y)$ that has maximal entropy, i.e., the one maximizing the *entropy functional*

$$H(f) = -\sum_{i=1}^{m} f_i \log f_i.$$

## 3. Comparing different strategies.

It is clear that the strategy $U$ is supposed to work "well". In an ideal situation it would give exact solution $U(f) = S(f)$ for any problem instance $f$. This is however usually impossible since, due to incomplete and/or noisy information, there are many elements $f_1$ sharing the same information $y$ and having different solutions $S(f_1)$. For instance, it is usually not possible to determine for sure if a patient is ill knowing only his name, or based only on visual investigation. Similarly, it is usually impossible to evaluate the integral based only on information that the integrand takes zero at $0$, $0.5$, and $1$. Existence of noise makes the situation even more difficult. In the above sense, the problem is *ill-posed*, and we are in an uncomfortable situation where we have to choose one "bad" strategy among many other "bad" strategies. How should we proceed? What strategy should we choose to approximate $S(f)$? One way to go is to pick a strategy that "seems to work well" and check on some examples if it "really works well". This *intuitive* or *heuristic* approach (although sometimes met in practice and often connected with some rational thinking) is not what we want to propose. For we aim in developing a general, rigorous theory, based on strictly defined mathematical components. A rigorous approach is to define something that will enable us to compare different strategies, and to select the best one.

We formally proceed as follows. Let $\mathcal{U}$ be a class of *admissible* strategies. (The restrictions on using strategies may have different sources, including limitations in information available or in computation capabilities.) We define on $\mathcal{U} \times \mathcal{U}$ a relation " $\prec$ " which makes $\mathcal{U}$ a *partially ordered* set. We say that a strategy $U_1 \in \mathcal{U}$ is *not worse* than $U_2 \in \mathcal{U}$ (or that $U_2$ is *not better* than $U_1$) iff

$$U_1 \prec U_2.$$

(Note that partial ordering means that not always two strategies can be compared.) We shall say that a strategy $U^*$ is *optimal* iff $U^* \prec U, \ \forall U \in \mathcal{U}$.

This is a very general and universal scheme. To be more specific, we now give some examples. For simplicity, we consider below only deterministic strategies.

**Example 4** In classical numerical analysis, one usually judges about a strategy by looking at how fast the successive approximations converge to the solution as the number of samples used goes to infinity. One introduces the notion of *exponent*. For a strategy $U = \{U_i\}$, the exponent is defined as the largest $\alpha$ (or the supremum of) such that for any "sufficiently smooth" (i.e., as many times differentiable as you please) function $f$ the error

$$|S(f) - U_{k_n}(f)| \leq K(f) \cdot n^{-\alpha}.$$

Here $K(f)$ is independent of $n$, and $k_n$ is the smallest index among such $i$ that $U_i$ uses at most $n$ samples. Then for two strategies we have $\ U^1 \prec U^2\ $ iff the corresponding exponents $\alpha_1 \geq \alpha_2$.

Consider, for instance, the integration problem of Example 1. For the trapezoidal rule we have

$$\int_0^1 f(x)\,dx \ = \ U_n^T(f) \ - \ \frac{1}{12n^2} f^{(2)}(\xi_1),$$

while for the Simpson rule we have

$$\int_0^1 f(x)\,dx \ = \ U_n^S(f) \ - \ \frac{1}{2280n^4} f^{(4)}(\xi_2),$$

where $\xi_1, \xi_2$ are some points in $[0, 1]$. Hence $U^T$ converges with exponent 2 and $U^S$ converges with exponent 4. Then $\ U^S \prec U^T$.

**Example 5** Another useful way of comparing strategies relies on introducing an error $\{e_i(U)\}_{i \in I}$. Then $U^1 \prec U^2$, e.g., iff $e_i(U^1) \leq e_i(U^2)$, $\forall i \in I$, or iff there is a $K$ independent of $i$ such that for all $i \in I$ we have $e(U_i^1) \leq K\, e(U_i^2)$. The error can be defined in different ways. One first introduces the notion of distance $dist(g_1, g_2)$ between elements in $G$ (If $F$ is a linear space then the distance is usually induced by a norm $||\cdot||$.) Then the error may be defined, e.g., as a *worst case* error

$$e_i^{wor}(U) = \sup\, \{\, dist\big(S(f), U_i(f)\big) : \; f \in F_0 \,\},$$

where $F_0$ is a class of problem instances included in $F$, or as an *average case* error as

$$e_i^{ave}(U) = \int_F dist\big(S(f), U_i(f)\big)\, \mu(df),$$

where $\mu$ is a probability measure on $F$ (or on a subset $F_0 \in F$), etc.

To be more specific, for the integration problem we have

$$dist\big(S(f), U_i(f)\big) = |S(f) - U_i(f)|.$$

In the worst case setting, one may take

$$F_0 = \{\, f \in C^r([0,1]) : \; ||f^{(k)}||_\infty \leq 1,\; 0 \leq k \leq r \,\}. \tag{4}$$

In the average case setting, one may assume that $\mu$ is an $r$-fold Wiener measure. (Recall that $\mu$ is for $r = 0$ the classical Wiener measure, or Brownian motion, characterized uniquely by the equality $\int_F f(s)f(t)\, \mu(df) = \min(s,t).$ )

**Example 6** In computational complexity theories, as *information-based complexity* or *theoretical computer science*, it is important to know the cost of obtaining an approximation with a given error $\varepsilon$. In this case, one first introduces a *computational model*. Then, based on this model, one defines a notion of *cost* of evaluating an approximation, and *complexity* of a strategy. Strategies are compared with respect to the complexity of obtaining an $\varepsilon$-approximation. More precisely, we have

$$\mathrm{comp}\,(U, \varepsilon) = \inf\, \{\, \mathrm{cost}\,(U_i) : \; e(U_i) \leq \varepsilon,\, i \in I \,\}.$$

We may define, e.g., $U^1 \prec U^2$ iff

$$\limsup_{\varepsilon \to 0^+} \frac{\mathrm{comp}\,(U^1, \varepsilon)}{\mathrm{comp}\,(U^2, \varepsilon)} < \infty.$$

Consider, for instance, the integration problem of Example 1. Let the error be the worst case error over the set $F_0$ defined as in (4) with $r = 4$. Let $\mathrm{cost}\,(U_i)$ be the number of samples used by $U_i$. Then, for the trapezoidal rule we have that $\mathrm{comp}\,(U^T, \varepsilon) \asymp \varepsilon^{1/2}$, while for Simpson rule $\mathrm{comp}\,(U^S, \varepsilon) \asymp \epsilon^{1/4}$. As a consequence, $U^S \prec U^T$.

**Example 7** Many existing techniques use what can be called *optimization functionals*. We already gave two examples of such functionals: the penalty functional of Example 2, and the entropy functional of Example 3. In such cases, one compares different strategies by comparing the corresponding values of the optimization functional. Note that the error (as defined in Example 5) can be also viewed as a special optimization functional.

## 4. Incorporating a priori information.

There is a number of possibilities of defining the relation " $\prec$ ", and examples of the previous section are just some specific cases. How to practically compare strategies (i.e., how to construct " $\prec$ ") is a very delicate question and we are going to discuss this point now. In any case, however, we have to be aware that we always have optimality with respect to some criterion. If the criterion (relation " $\prec$ ") changes, another strategy may turn out to be optimal.

First of all, we notice that there are in general no strategies that are universally good. If the problem is not trivial then, for any two "reasonable" strategies $U_1$ and $U_2$ (with the index set $I$ being a singleton, for simplicity), we can find a set of $f$'s for which $U_1(f)$ is closer (or even equal) to $S(f)$ than $U_2$, but there is also another set of $f$'s for which $U_2(f)$ is closer to $S(f)$ than $U_1(f)$. At first sight, this observation (which is just the consequence of the fact that information is only partial and/or noisy) may lead to the pessimistic conclusion that any attempts to construct a reasonable relation " $\prec$ " are hopeless, since too many strategies cannot be compared with each other. On the other hand, such a conclusion would contradict all the practical computations, where many existing strategies for solving different kind of problems are known to be extremely powerful. The point is that those "powerful strategies" are tested and then used only for problem instances $f$ possessing some additional, particular properties. These properties are in a natural way incorporated into the definition of the problem and influence the process of designing a strategy. (Indeed, it is much easier to recover a picture from only parts of it if we know that this is the picture of a chair. It is easier to guess whether a patient is ill if we restrict considerations only to patients who are old and smoked in the past, etc.)

It often happens that we know much more about the problem instance $f$ than just the obvious fact that it is in the domain $F$ of the mapping $S$. For example, if $f$ is a function representing an image, then we may know that $f$ is in a sense "smooth", or that it has sudden jumps or sharp edges, etc., depending on what kind of image $f$ is supposed to represent. If $f$ is a distribution that we want to recover then we may know that uniform distributions are more likely than non uniform ones. In the case of noisy information, we may also know something about the noise $\mathfrak{n}$, e.g., that it is bounded, Gaussian, etc. Such statements can be formally described as, e.g., $f \in F_0 \subset F$ or $\mathfrak{n} \in Y_f \subset Y$, and/or that these are distributed (or are stochastic processes) according to known probability measures, $f \sim \mu$ or $\mathfrak{n} \sim \pi_f$, etc. This kind of information will be called *a priori information* about $f$, as opposed to the information $y$ about $f$, introduced earlier, which comes from additional observations (measurements, computations) on $f$. The latter will be called, in contrast, *a posteriori information*.

We now give one simple example of how a priori information can be incorporated into the definition of the problem and, consequently, how it can imply the definition of " $\prec$ ".

**Example 8** Suppose we want to know the value of a real parameter $f$ based on information $y = f + \mathfrak{n}$, where $\mathfrak{n}$ is some noise. In this case, it is natural to put $U(f, \mathfrak{n}) = A_1(y) = y$. Does this strategy work well? How to compare it with other strategies? If we do not put any additional assumptions on $f$ and/or $\mathfrak{n}$ then we are hopeless. Hence we assume that for any $f$ the noise $\mathfrak{n}$ has normal distribution with mean 0 and variance $\sigma^2 > 0$, and we allow any measurable strategies (algorithms) $A : \mathbb{R} \to \mathbb{R}$. Then we may define the error of any $A$ as

$$e(A) = \sup_{f \in \mathbb{R}} e(A, f), \tag{5}$$

where $e(A, f)$ is the (average) error for particular $f$,

$$e(A, f) = \Big(\frac{1}{\sqrt{2\pi\sigma^2}} \int_{\mathbb{R}} |f - A(f + \mathfrak{n})|^2 \exp(-\mathfrak{n}^2/(2\sigma^2))\, d\mathfrak{n}\Big)^{1/2}.$$

With such an error, it turns out that the algorithm $A_1$ is optimal indeed, and its error equals just $\sigma$. Hence the algorithm works very well provided the noise is zero mean normal, and $\sigma$ is "small". If, however, we knew a priori that $|f| \leq \tau$, the algorithm would not be optimal since, as a consequence of noise, we may have $|A_1(y)| > \tau$. A better algorithm would be $A(y) = y$ for $|y| \leq \tau$, $A(y) = -\tau$ for $y < -\tau$, and $A(y) = \tau$ for $y > \tau$. However, to make this algorithm better not only conceptually, but also formally, we have to change the definition of error to

$$e(A) = \sup_{f \in [-\tau, \tau]} e(A, f),$$

so that it now corresponds more to the assumption (our a priori knowledge) on $f$.

Another kind of assumption about $f$ could be that it is a random variable, e.g., zero mean normal with variance $\tau^2 > 0$. In such a case, it is natural to define the error as the average error

$$e(A) = \Big(\frac{1}{\sqrt{2\pi\tau^2}} \int_{\mathbb{R}} e(f, A)^2 \exp(-f^2/(2\tau^2))\, df\Big)^{1/2}.$$

The algorithm $A_1(y) = y$ is again not optimal. Indeed, it does not incorporate the a priori information that it is more probable that $f$ is closer to 0 than to any other number. An algorithm that shifts the information $y$ a little towards zero should be better. Formally, we still have that $e(A_1) = \sigma$, while the minimal error that can be achieved in this case equals

$$e_{min} = \sigma \cdot \sqrt{\frac{1}{1 + \sigma^2/\tau^2}},$$

and is obtained by the algorithm

$$A_2(y) = \left(1 + \frac{\sigma^2}{\tau^2}\right) \cdot y.$$

These formulas illustrate very well how different a priori assumptions (here the values of $\delta$ and $\tau$) influence the choice of algorithm (strategy) and its error. We note that the initial definition (5) can be viewed as the case where we do not have any a priori information about $f$, but only about the noise. This is the limiting case when $\tau \to +\infty$, i.e., when the distribution of $f$'s becomes more and more "uniform".

## 5. Optimization functionals.

We saw that, when solving a problem, two kinds of information has to be taken into account: a priori and a posteriori. While the a posteriori information has a rather objective character (this is just our numerical data $y = N(f,\mathfrak{n})$ about $f$), a priori information has in most cases a subjective character. Indeed, initially it is described in a very general way (e.g., that the object is "smooth", that "most of $f$'s" are in a specific region, that the distribution of $f$'s is "uniform", etc.), and only then one tries to find a mathematical formulation for such information in order to use it to compare different strategies and, eventually, to choose the best one. We stress that, in theoretical considerations, we frequently assume even more about $f$ (and possibly also about noise $\mathfrak{n}$ if it exists) than our a priori information indicates. The reason is simple: some problems are so difficult that it is impossible to say something reasonable about different strategies of solving it without putting some additional, sometimes maybe unrealistic, assumptions. A good example is the problem of solving a complicated partial differential equation (PDE). When analyzing algorithms for solving PDE's one usually assumes that the coefficients or even the solution itself have some degree of smoothness, even though everyone is aware that such assumptions are, as a rule, not met in reality. We want to stress this, although from the point of view of pure mathematical formulation it is not important at all whether the a priori information is *given* in a natural way, or it is *assumed*.

Even though the question what can and what cannot be assumed as a priori information belongs to philosophy rather than to mathematics, it is impossible to ignore it. For any specific answer immediately implies what mathematical model and tools will be used to solve the problem. The differences in a priori assumptions may even lead to divisions among researchers dealing with seemingly the same problems, as happened, e.g., with Bayesian and non-Bayesian statistics. Here is another example.

**Example 9** In *information-based complexity* theory, one compares strategies using the concepts of error and cost. For simplicity, we concentrate on the error only, namely the worst case error. Supposing that $F$ and $G$ are linear spaces, the (global) *worst case error* of an algorithm $A$ using (deterministic) information $y = N(f)$ is defined as

$$e(A) = \sup_{f \in F_0} ||S(f) - A(N(f))||.$$

Here $||\cdot||$ is a norm on $G$, and $F_0$ is a subset of $F$, e.g., the unit ball with respect to some seminorm $||\cdot||_F$,

$$F_0 = \{\, f \in F : \ ||f||_F \leq r \,\}.$$

A good thing about this model is that the error is always well defined and any two strategies, even those using different information, can be easily compared. However, for some researches, as classical *numerical analysts*, it is difficult to accept this definition, because it explicitly states with certainty that $f$ is in a specified ball, and this information is usually not available. For instance, even if we know that $f$ is a function with bounded second derivative, we usually do not know the bound itself. That is why in classical numerical analysis one compares strategies using the concept of *speed of convergence* rather than the (global) error, as explained in Example 4. (Surprisingly enough, the two approaches are not so different as it may seem at first sight, as will be explained later.)

If we have a priori information that $f \in F_0$, or that $f \sim \mu$, then one may in a natural way define the relation " $\prec$ " using, e.g., the concept of the worst case or average case error, correspondingly, as in Example 5. As noticed in Example 9, a practical difficulty is, however, that even though we know a priori that $f$ possesses a property, say $(P)$, it is usually impossible to give or even estimate a quantitative value of $(P)$. For instance, even if we know that $f$ is a function with bounded second derivative, we usually do not know the bound itself. Or, even if we know that $f$ has zero mean Gaussian distribution, we usually do not know the covariance operator exactly. How should we proceed if we do not want to accept any additional assumptions about $f$? A commonly applied and quite natural approach in such situations relies on using *optimization functionals*.

Suppose that a posteriori information about $f$ is fixed. That is, we formally consider only strategies $U = (A, N)$ with fixed $N$. Let us first assume, for simplicity, that we are in the deterministic case. In order to find a "good" approximation for $S(f)$ based on information $y = N(f)$, we proceed as follows. For the a priori property $(P)$, we define a corresponding functional

$$\psi : F \to [0, +\infty] = [0, +\infty) \cup \{+\infty\}.$$

The value $\psi(f)$ gives us quantitative information about to what extent $(P)$ is satisfied by $f$. We adopt the convention that the less $\psi(f)$ the more "$(P)$ is present in $f$." Then we put $A^O(y) = S(\mathfrak{s}_y)$, where

$$\mathfrak{s}_y = \arg\min \{\, \psi(h) : \ h \in F, \ N(h) = y \,\}. \tag{6}$$

(We assume, for simplicity, that the minimum above is attained.) Hence, we pick as approximation the image of the element which is consistent with information $y$ and satisfies $(P)$ "as much as possible". (We note that this approach formally corresponds to the relation " $\prec$ " defined as, e.g., $(A_1, N) \prec (A_2, N)$ iff for any information $y$ there is $f_y$ such that $A_1(y) = S(f_y)$, and $A_2(y) \neq S(h)$ for any $h \in N^{-1}(y)$ with $\psi(h) < \psi(f_y)$ $N(h) = y$.)

**Example 10** One of the most striking examples of using optimization functionals is provided by classical *splines*. That is, consider the problem of recovering a function $f : [a, b] \to \mathbb{R}$ from information $y = (y_1, \ldots, y_n)$ where $y_i = f(t_i)$, $0 \leq i \leq n$, and $a = t_0 < t_1 < \cdots < t_n = b$. The a priori assumption is that the function $f$ is "smooth". This assumption can be incorporated to the mathematical formulation of the problem by introducing the functional

$$\psi(f) = \int_a^b \left(f^{(2)}(t)\right)^2 df$$

(and $\psi(f) = +\infty$ if the integral is not well defined), which roughly represents the "amount of smoothness in $f$". The solution of (6) is then the natural cubic spline interpolating the data $y$.

The situation becomes more complicated if information is corrupted by some noise, $y = N(f) + \mathfrak{n}$. In this case, using the same optimization functional would not be appropriate. Indeed, the procedure would lead to exact interpolation of data. However, because of the noise, this data may be "rough" and hence the minimization of $\psi(f)$ could result in $\mathfrak{s}_y$ which only "weakly" satisfies the a priori assumption $(P)$. Hence we have to trade between interpolating data and maximizing $(P)$. A natural way to go is as follows. We first define two functionals, $\psi : F \to [0, +\infty]$ and $\phi_y : F \to [0, +\infty]$, $y \in Y$, which represent the property $(P)$ and fitness to the data $y$, correspondingly. Then we combine these functionals in some way to obtain a functional $\psi_y : F \to \mathbb{R}_\infty$ which is to be minimized over the whole space $F$..

**Example 11** We generalize the problem of Example 10 to the case of noisy information, i.e., $y_i = f(t_i) + \mathfrak{n}_i$, $0 \leq i \leq n$. We define, e.g., $\phi_y(f) = n^{-1} \sum_{i=0}^{n} (y_i - f(t_i))^2$, and

$$\psi_y(f) = \int_a^b \left(f^{(2)}(t)\right)^2 df + \frac{\lambda}{n} \cdot \sum_{i=0}^{n} (y_i - f(t_i))^2.$$

Here $\lambda$ is a parameter which controls the tradeoff between the smoothness of $f$ and fitness to the data $y$; the more $\lambda$, the more we trust the data. This kind of optimization functionals is widely used in statistical estimation. How to choose $\lambda$ is a separate problem and it can be solved again using some a priori assumptions or by optimizing another functional. The latter approach is represented by, e.g., the well known *cross validation* techniques.

# 5. Equivalence of different techniques.

We presented a couple of techniques of incorporating a priori information into the mathematical formulation of the problem. We now show some rather surprising results to the effect that these techniques are not as different as may seem at first sight, and that in many cases they eventually lead to similar strategies of solving the problem.

### 5.1. Optimization and worst case approach.

Suppose, as in Section 4, that information $N$ is fixed. Furthermore, we have functionals $\psi, \phi_y : F \to [0, +\infty]$, $y \in Y$, which measure amount of property $(P)$ in $f$, and fitness of $f$ to the data $y$, correspondingly. Using the worst case approach, we may define the error of an algorithm $A$ as

$$e(A, N) = \sup \{ dist(S(f), A(y)) : f \in F, y \in Y, \text{ s.t. } \psi(f) \leq r, \phi_y(f) \leq \delta \},$$

where $dist$ is a metric in $G$. Note that we have two parameters, $r$ and $\delta$, which have to be chosen according to our belief (or knowledge) about the properties of $f$ and the noise level in the information. The case $\delta = 0$ will be interpreted as exact (non-noisy) information.

In this case, the minimal error that can be achieved equals

$$r(N) = \inf_A e(A, N) = \sup_{y \in Y} rad\big(S(F_y)\big),$$

where $F_y = \{f \in F : \psi(f) \leq r, \phi_y(f) \leq \delta\}$ is the set of $f$'s consistent with our a priori knowledge and information $y$, and $rad(\,\cdot\,)$ is the (Chebyshev) radius of a set. An optimal algorithm gives as $A^*(y)$ the center of $S(F_y)$ (if it exists).

As the center is sometimes difficult to construct, one often uses the *interpolation* approach to construct almost optimal algorithms. Specifically, one puts

$$A^I(y) = S(f_y)$$

where $f_y$ is any element interpolating our a priori and a posteriori knowledge, i.e., such that $f_y \in F_y$. Then the error of $A_I$ is at most twice worse than the minimal error. Indeed, since $A(y) = S(f_y) \in S(F_y)$, we have for any $f \in F_y$ that

$$dist\big(S(f), A^I(y)\big) \leq diam\big(S(F_y)\big) \leq 2 \cdot rad\big(S(F_y)\big)$$

($diam(\,\cdot\,)$ is the diameter), which gives $e(A^I, N) \leq 2 \cdot r(N)$.

Let us now see what gives us the optimization functionals approach. We define

$$\psi_y(f) = \max\big(\psi(f), \lambda \phi_y(f)\big),$$

where $\lambda$ is a parameter. To include also the exact information case, we also allow $\lambda = \infty$ (with convention $\infty \cdot 0 = 0$). Let $f \in \{h \in F : \psi_y(f) < \infty\}$, and let $y$ be information about $f$. Let $M < \infty$ be such that $\psi(f) \leq M$ and $\phi_y(f) \leq M/\lambda$. The optimization approach gives $A(y) = S(\mathfrak{s}_y)$ where $\mathfrak{s}_y$ minimizes $\psi_y(\,\cdot\,)$ (assuming, for simplicity, that the minimum is attained). We have $\psi_y(\mathfrak{s}_y) \leq \psi_y(f) \leq M$, which implies that $\psi(\mathfrak{s}_y) \leq M$ and $\phi_y(\mathfrak{s}_y) \leq M/\lambda$. Hence $A^O$ is nothing but the interpolatory algorithm provided our a priori knowledge is $\psi(f) \leq M$ and $\phi_y(f) \leq M/\lambda$. We conclude that the optimization approach leads to an algorithm which is optimal (within a factor of 2) in the worst case setting with respect to any "balls" $\psi(f) \leq r$ and $\phi_y(f) \leq \delta$ whose "radii" satisfy

$$\delta/r = \lambda^{-1}.$$

Note that this relation is especially striking for exact information, since then we just minimize $\psi(h)$ over $h$ such that $\psi_y(h) = 0$. For noisy information, we have to know the ratio $\delta/r$. We already mentioned this difficulty in Example 11.

### 5.2. Optimization and asymptotic approach.

A relation between the optimization approach and the asymptotic approach described in Example 4 is less obvious. Therefore we present it only in its simplest form. We assume exact information. A strategy is a sequence of algorithms $\{A_n(N_n(\,\cdot\,))\}_{n \geq 0}$, where

$$N_n(f) = \bigl(L_1(f), L_2(f), \ldots, L_n(f)\bigr) \in \mathbb{R}^n.$$

We stress that information is here *nested*, i.e., for each $n$, information $N_n$ consists of the $n$ first functionals of a preselected infinite sequence $\{L_i\}_{i \geq 1}$. We also need more specific assumptions. We assume that $F$ is a Banach space with a norm $\|\cdot\|_F$ which represents the property $(P)$, and $G$ is a normed space with a norm $\|\cdot\|$, so that the distance in $G$ can be measured in this norm. The functionals forming information are continuous linear, and $S : F \to G$ is also a continuous linear mapping.

Let $A_n^O$ be the algorithm that uses information $N_n(f)$ and results from optimization of the functional $\psi(f) = \|f\|_F$. Then $A^O$ is interpolatory and from Section 5.1 we obtain

$$\|S(f) - A_n^O(N_n(f))\| \leq \sup\bigl\{\|S(h_1) - S(h_2)\| : \|h_j\|_F \leq \|f\|_F,\; N_n(h_j) = N_n(f)\bigr\}.$$

Now, some calculations using linearity of $S$ and $N_n$ give

$$\|S(f) - A_n^O(N_n(f))\| \leq 2\|f\|_F \sup\bigl\{\|S(h)\| : \|h\|_F \leq 1,\; N_n(h) = 0\bigr\} = 2\|f\|_F\, r(N_n),$$

where $r(N_n)$ is the minimal worst case error of the $n$th information with respect to the unit ball in $F$. Thus we obtained that, for given $f \in F$, the error of $A_n^O$ tends to zero at least as fast as the sequence of the worst case errors.

It turns out that this speed of convergence cannot be essentially improved; namely, we have the following theorem. Let $\{\eta_n\}_{n \geq 0}$ be any positive sequence converging to zero. Let $\{A_n(N_n(\,\cdot\,))\}_{n \geq 0}$ be any strategy. Then for any nontrivial ball $B \subset F$ there exists $f \in B$ such that the sequence of errors $\|S(f) - A_n(N_n(f))\|$ does not converge to zero faster than the sequence $\{\eta_n r(N_n)\}$.

The above correspondence has generalizations to noisy information and also to some nonlinear problems as, e.g., solving ordinary differential equations (ODE's).